# Fifth system
# for
# general-purpose connectionist computation

Anthony Di Franco
Berkeley, CA

## Summary

To date, work on formalizing connectionist computation in a way that is at least Turing-complete has focused on recurrent architectures and developed equivalences to Turing machines or similar super-Turing models, which are of more theoretical than practical significance.[1] We instead develop connectionist computation within the framework of information propagation networks[2] extended with unbounded recursion, which is related to constraint logic programming[3] and is more declarative than the semantics typically used in practical programming, but is still formally known to be Turing-complete.

This approach yields contributions to the theory and practice of both connectionist computation and programming languages. Connectionist computations are carried out in a way that lets them communicate with, and be understood and interrogated directly in terms of the high-level semantics of a general-purpose programming language. Meanwhile, difficult (unbounded-dimension, NP-hard) search problems in programming that have previously been left to the programmer to solve in a heuristic, domain-specific way are solved uniformly *a priori* in a way that approximately achieves information-theoretic limits on performance.

Connectionist augmentations for representing high-dimensional spaces concisely with information-theoretic autoencoders are juxtaposed with program structures along the boundaries of program recursion, then composed in a hierarchy along branches of the recursion. The compression that results permits communicating information in arbitrarily deep recursions at costs proportional to the logarithm of the recursion depth and information to be communicated, and permits identification of similar subtrees of the recursion. This forms the basis of a general and efficient means of solving search, planning, scheduling, and optimization problems guided by learned connectionist representations of the search space. It also rediscovers the essence of biologically-inspired hierarchical models of information processing in cortex, establishing the relevance of the same concepts to theory of computation in terms of information-theoretic approximation. Results generalize work on strictly linear and harmonic state-space compression in POMDPs.[4]

# Additional Detail

The crux of the approach is to take advantage of the information propagation framework's ability to abstract away from control issues in working out the implications of new information, and to learn efficient representations of search spaces without making distributional assumptions by using autoencoders that work in strictly information-theoretic terms and compose them in a way that closely mirrors recursive composition in programming. The resulting composition trivially retains the Turing-complete or equivalently Chomsky-unrestricted computational power of the base system based only on local propagation of information.

The previously mentioned autoencoder hierarchy is built up recursively, first, by instantiating an autoencoder alongside each of the sets of information states within each recursive definition in the program, then by hierarchically and recursively composing autoencoders that are adjacent in the information graph of the program by instantiating another autoencoder that communicates between the representations of each adjacent node. Self-recursive nodes, as well as all nodes instatiated from the same definition, all inform each other, which is the basis for identification of common subtrees. The picture that emerges has tree structure in two dimensions – first, in the program's definitions, reflected by the information graph at any point in evaluation, and second a tree that sits upon each recursive information propagation path starting at the root of the definition graph, and handles the long-range communication of information.

A query identifies certain nodes for scrutiny, and these nodes serve as roots in storage management and the source of approximation bounds to be satisfied by the overall information propagation process. Indexing the recursion with information constrained by a total order, a way of modeling local constraints of sequentiality and time is obtained in the style of temporal logics. This is the basis of expressing constraint structures that expresses the relationship of present states and actions to temporally local future states, action options, and rewards, and thus for solving planning and scheduling problems and (PO)MDPs via the automatically-arranged propagation of information across larger scales in time and space.

The preferred autoencoder known so far for this system is LOCOCODE,[5] which like ICA[6] uses a purely information-theoretic objective, but also does not require the dimensionality of the approximation to be specified, and separates the mappings into and out of the representation space and allows them to be arbitrary functions.

One can adopt and freely intermix and interpolate between, on the one hand, a style that emphasizes *a priori* knowledge of mutual constraints expressed as a constraint logic program, and uses the connectionist network augmentation only to facilitate non-local inference, and, on the other hand, a style that merely recursively decomposes raw data and uses the connectionist augmentations to the network to inductively discover the entire representation of regularities in the domain. When they are intermixed, the one kind of information informs the other without special effort.

Additionally, as a desirable side-effect of basing computation on propagation of partial information, the typing and evaluation of the program are unified under the propagation of partial information about values, eliminating type/value and compile/runtime distinctions in programming that introduce restrictions on expressive power and complications in implementation where they appear elsewhere.

---

5   Hochreiter, Sepp and Schmidhuber, Juergen. LOCOCODE. IDSIA Technical Report FKI-222-97, 1997.
6   Bell A.J. and Sejnowski T.J. 1995. An information maximisation approach to blind separation and blind deconvolution, *Neural Computation,* 7, 6, 1129-1159